\documentclass[letterpaper, 10 pt, conference]{ieeeconf}  

\usepackage{amsmath}
\usepackage{color}
\usepackage{float}
\usepackage{graphicx}
\usepackage{caption}
\usepackage{subcaption}
\usepackage{stfloats}
\usepackage[colorlinks=True,linkcolor=red,citecolor=red]{hyperref}
\usepackage{url}
\usepackage{placeins}
\usepackage{afterpage}
\usepackage{makecell}
\usepackage{diagbox}

\usepackage{wrapfig}

\IEEEoverridecommandlockouts                              
\title{\LARGE \bf
MuJoCo MPC for Humanoid Control: Evaluation on HumanoidBench}

\author{Moritz Meser\textsuperscript{1,2}, Aditya Bhatt\textsuperscript{2}, Boris Belousov\textsuperscript{2} and Jan Peters\textsuperscript{1,2,3,4}
\thanks{\textsuperscript{1}Intelligent Autonomous Systems Lab, Department of Computer Science, TU Darmstadt, Germany, {\tt\footnotesize moritz.meser@stud.tu-darmstadt.de}}
\thanks{\textsuperscript{2}German Research Center for AI (DFKI)}
\thanks{\textsuperscript{3}Centre for Cognitive Science, Technical University of Darmstadt}
\thanks{\textsuperscript{4}Hessian Center for Artificial Intelligence (Hessian.AI), Darmstadt}
\thanks{We thank Hessisches Ministerium für Wissenschaft \& Kunst for the DFKI grant and ``The Adaptive Mind'' grant.}
}

\begin{document}

\maketitle
\thispagestyle{empty}
\pagestyle{empty}

\begin{abstract}
We tackle the recently introduced benchmark for whole-body humanoid control \texttt{HumanoidBench}~\cite{sferrazza2024humanoidbench} using MuJoCo MPC~\cite{howell2022predictive}.
We find that sparse reward functions of \texttt{HumanoidBench} yield undesirable and unrealistic behaviors when optimized; therefore, we propose a set of regularization terms that stabilize the robot behavior across tasks.
Current evaluations on a subset of tasks demonstrate that our proposed reward function allows achieving the highest \texttt{HumanoidBench} scores while maintaining realistic posture and smooth control signals.
Our code is publicly available and will become a part of MuJoCo MPC\footnote[5]{\scriptsize \url{https://github.com/google-deepmind/mujoco\_mpc/pull/328}},
enabling rapid prototyping of robot behaviors.

\end{abstract}

\section{Introduction and Related Work}
Designing dynamic behaviors for humanoid robots is a challenging problem~\cite{hansen2024hierarchical,he2024learning}.
One promising approach is to craft task-specific reward functions and train policies via Reinforcement Learning (RL).
To this end, \cite{sferrazza2024humanoidbench} proposed \texttt{HumanoidBench}, a suite of 27 tasks for Unitree H1 robot based on MuJoCo~\cite{todorov2012mujoco}, with 12 tasks assessing locomotion abilities and 15 tasks evaluating object manipulation skills.
In~\cite{sferrazza2024humanoidbench}, performances of four popular RL algorithms are reported: DreamerV3~\cite{hafner2023mastering}, TD-MPC2~\cite{hansen2023td}, SAC~\cite{haarnoja2018soft},  PPO~\cite{schulman2017proximal}.
However, none of these methods is able to solve the tasks satisfactorily, producing
jittery motions ill-suited for execution on real hardware.
Therefore, we evaluate an alternative approach to humanoid behavior synthesis that exploits the power of simulation: Model Predictive Control (MPC).

MPC does not require training; at each time-step, it selects the optimal action based on a multi-step lookahead search, and repeats this procedure at the next time-step in a receding-horizon manner~\cite{moerland2023model}.
MPC is a mature paradigm that admits various planning algorithms for this search, including iLQG~\cite{tassa2012synthesis} and even randomly sampled action sequences~\cite{howell2022predictive}.
We apply MPC to a subset of the \texttt{HumanoidBench} locomotion and manipulation tasks using MuJoCo MPC (MJPC)~\cite{howell2022predictive}, which implements the aforementioned planners.

\section{MJPC with Shaped Rewards}
\begin{figure}[t]
    \centering
    \includegraphics[width=0.8\linewidth]{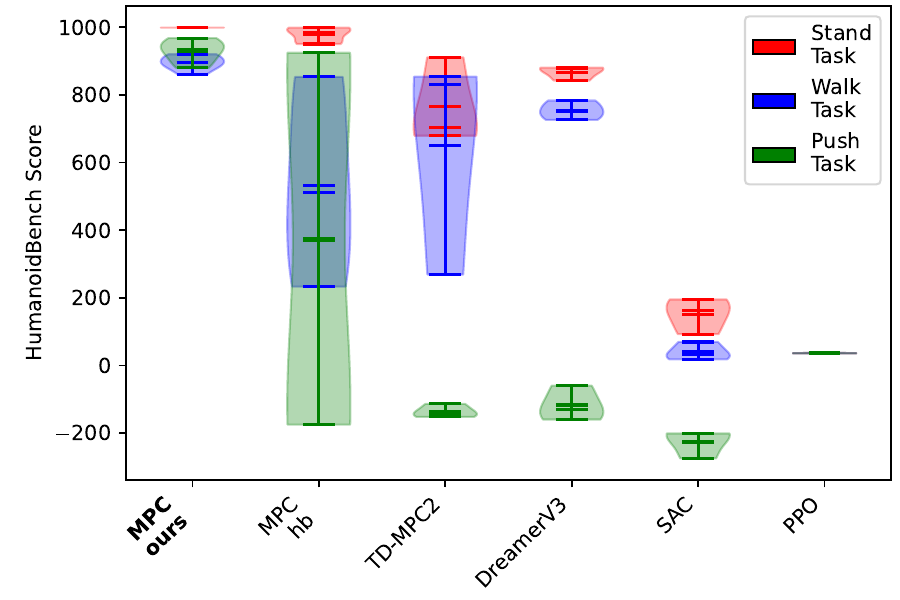}
    \caption{
    Performance comparison between the proposed controller \textit{MPC-ours} that leverages shaped reward functions and the baselines: \textit{MPC-hb}, which uses the \texttt{HumanoidBench} reward function, and the RL baselines.
    The $y$-axis shows the \texttt{HumanoidBench} score given by the sum of rewards over a trajectory, with maximum $1000$ for each task.
    For MPC methods, we employ the iLQG planner on the \texttt{Stand} and \texttt{Walk} tasks, and the Sampling planner on the \texttt{Push} task which involves a lot of contact interactions.
    Results from $6$ runs of each MPC method are reported.
    RL baseline results are imported from~\cite{howell2022predictive} where $3$ runs of each method are reported; we take the best policy from each run.
    \textit{MPC-ours} significantly outperforms the baselines across tasks.
    }
    \label{fig:plots_overall_ref}
    \vspace{-1em}
\end{figure}
\begin{figure*}[t]
  \centering
    \includegraphics[width=\textwidth]{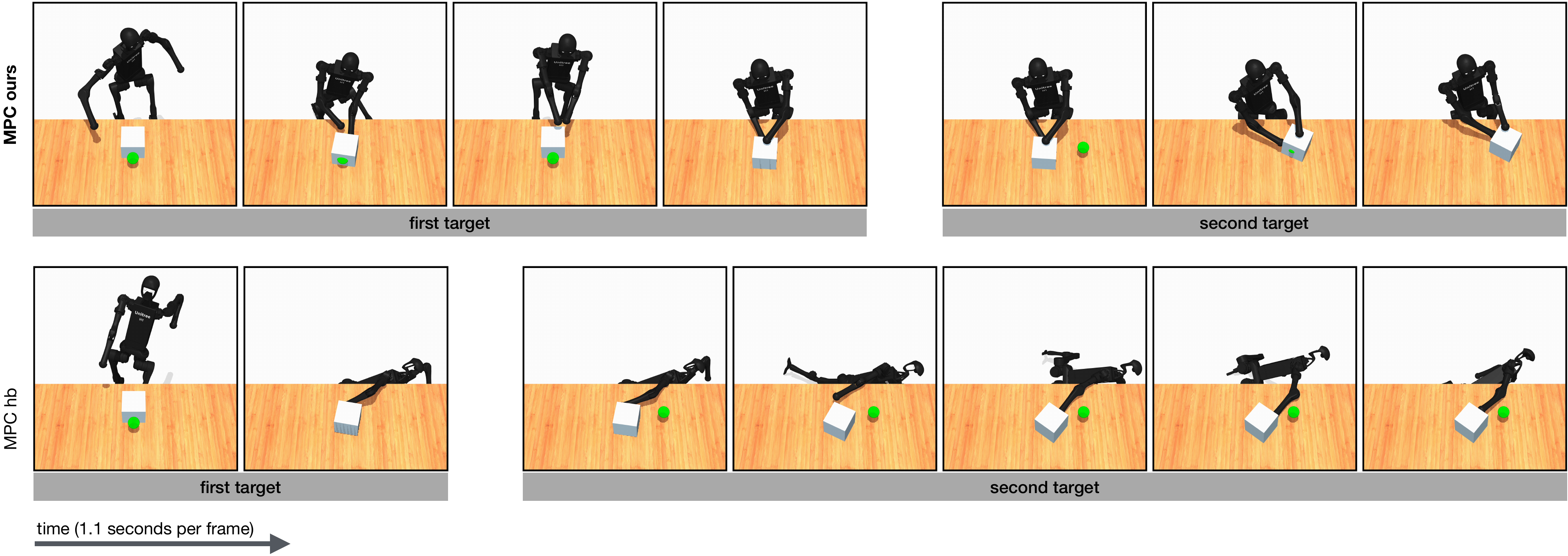}
    \caption{
    Behavior comparison between \textit{MPC-ours} (top row) and \textit{MPC-hb} (bottom row) on the \texttt{Push} task.
    Unlike our shaped reward, which encourages posture maintenance and balance, the \texttt{HumanoidBench} reward puts all emphasis on reaching the target box location as fast as possible, driving the robot into unrecoverable postures and thereby precluding further tasks.
    }
  \label{fig:snapshot_push_box}
  \vspace{-1em}
\end{figure*}

MPC repeatedly solves a finite-horizon optimal control problem $\min_{u_{0:T}} \sum_{t=0}^{T}{c(x_t, u_t)}$ where $c(x_t, u_t)$ is the instantaneous cost.
MJPC further decomposes the cost into a weighted sum 
$
c(x,u) = \sum_{i=0}^{M} w_i \cdot n_i (c_i(x,u))
$
where $c_i(x,u)$ are (signed) residuals, $n_i$ are twice-differentiable norm functions, and $w_i$ are tunable non-negative weights~\cite{howell2022predictive}.

The \texttt{HumanoidBench} tasks are formulated in terms of rewards $r_\text{hb}$ rather than costs.
Therefore, we apply a transformation $c_\text{hb}(x,u) = |r_{\text{max}} - r_{\text{hb}}|$ before passing it to MJPC, where $r_\text{max}$ is the maximum achievable instantaneous reward (set to $r_\text{max} = 1$) and the norm is approximated by \texttt{kSmoothAbsLoss} to ensure differentiability. 

We observed that $r_\text{hb}$ alone yields physically undesirable behaviors (Fig.~\ref{fig:snapshot_push_box}) and results in low rewards (Fig.~\ref{fig:walk_long_duration}).
Therefore, we introduced a number of shaping terms that i)~improve robot stability and ii) provide dense reward signal.

\textit{Stability-enhancing cost terms} penalize excessively strong movements and promote postural balance:
1) height of the robot's head from the ground;
2) height difference between the pelvis and the feet;
3) linear velocity of the center of mass (CoM);
4) `balance' (projection of CoM between the feet);
5) deviation from a `canonical' posture;
6) direction the robot is facing;
7) magnitude of the control signal.

\emph{Dense reward residuals}:
1) distance between an object and its target location;
2) distance between the left hand and the object;
3) distance between the right hand and the object.
Although some of these residuals are already encoded in the \texttt{HumanoidBench} reward, we find that providing them individually to the optimizer improves  performance (Fig.~\ref{fig:plots_overall_ref}).

\section{Extended Task Evaluation Protocol}
\begin{figure}[t]
    \centering
    \includegraphics[width=\linewidth]{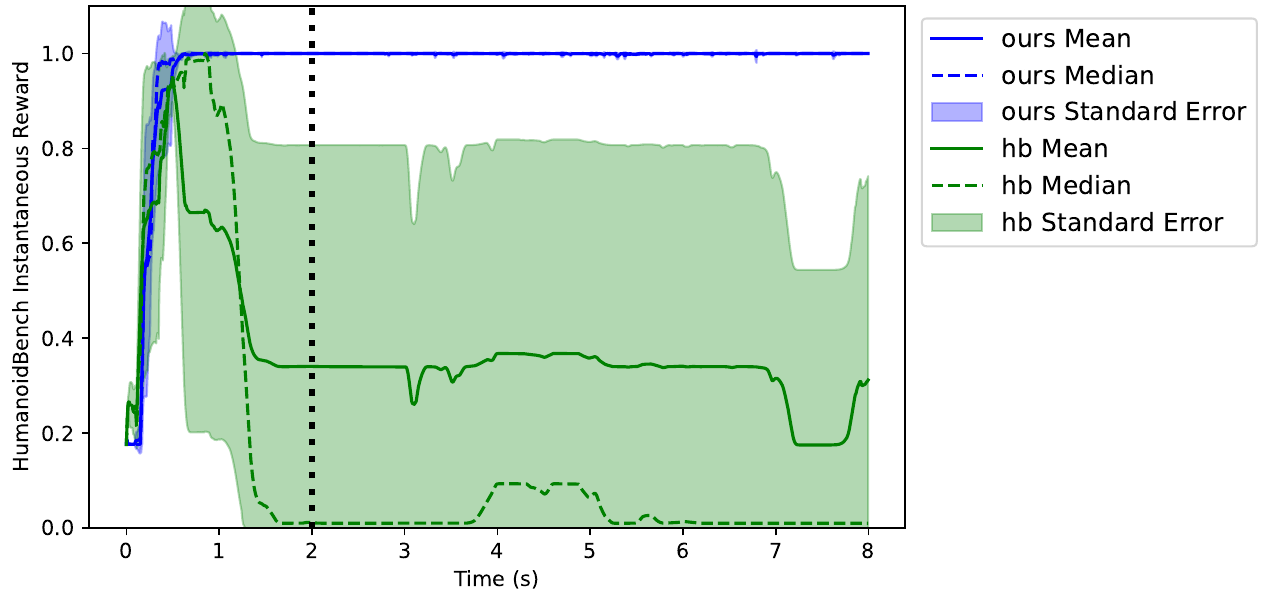}
    \caption{
    Influence of the episode length on evaluation scores shown on the \texttt{Walk} task over $6$ runs.
    An episode would normally stop at the $2$s mark (vertical dotted line), yielding a relatively high cumulative reward for \textit{MPC-hb} (green curve) thanks to good initial performance.
    However, when extended further, the median \texttt{HumanoidBench} reward drops almost to zero, while ours maintains the maximum value of $1$.
    }
    \label{fig:walk_long_duration}
    \vspace{-1.5em}
\end{figure}
Along with the pointed out uninformativeness of the reward function in \texttt{HumanoidBench}, we observe one further factor that impacts the generated behaviors and evaluations---namely, the \emph{episode length}.
The evaluation period
is very short---2 seconds for the \texttt{Walk} and \texttt{Stand} tasks; and for the \texttt{Push} task, the simulation terminates as soon as the box reaches its target location.
Such short episode lengths result in policies that do not maintain balance at the end of the episode, not expecting further tasks (Fig.~\ref{fig:snapshot_push_box}, bottom).
We provide a quantitative evaluation of the impact of the episode length in Fig.~\ref{fig:walk_long_duration} on the \texttt{Walk} task.
A qualitative evaluation is provided in Fig.~\ref{fig:snapshot_push_box}, where new targets are re-spawn, requiring the robot to perform the task repeatedly.
Based on these evaluations, we argue for longer episode lengths and for repeated tasks with changing goals.

\section{Analysis of Behaviors and Planners}
In this section, we analyze the generated behaviors in a quantitative manner, beyond only reporting the return value.
\begin{wraptable}{r}{0.6\linewidth}
    \vspace{-0.25em}
    \centering
    \caption{Control smoothness cost: average squared joint velocity.}
    \begin{tabular}{|c|c|c|}
    \hline
    \footnotesize{\textbf{Task}} & \footnotesize{\textbf{MPC-ours}} & \footnotesize{\textbf{MPC-hb}} \\
    \hline
    \footnotesize{\texttt{Walk}} & \footnotesize{\textbf{1.45} \(\pm\)} \(\text{\scriptsize 0.13}\) & \footnotesize{1.67 \(\pm\)} \(\text{\scriptsize 0.87}\) \\
    \hline
    \footnotesize{\texttt{Stand}} & \footnotesize{\textbf{0.01} \(\pm\)} \(\text{\scriptsize 0.00}\) & \footnotesize{1.49 \(\pm\)} \(\text{\scriptsize 0.23}\) \\
    \hline
    \footnotesize{\texttt{Push}} & \footnotesize{1.60 \(\pm\)} \(\text{\scriptsize 0.14}\) & \footnotesize{1.66 \(\pm\)} \(\text{\scriptsize 0.05}\) \\
    \hline
    \end{tabular}
    \label{smoothness_cost}
    \vspace{-0.75em}
\end{wraptable}
First, we evaluate the smoothness of trajectories (energy cost), as judged by the average squared joint velocity (Table~\ref{smoothness_cost}).
Our reward generates consistently smoother trajectories on the \texttt{Stand} and \texttt{Walk} tasks (while also achieving much higher reward---see Fig.~\ref{fig:walk_long_duration}).
On the \texttt{Push} task, the cost is comparable due to the short episode duration.
Analysis with respect to other criteria will be provided in an extended version of the article.

\begin{table}[h]
    \vspace{-0.75em}
    \centering
    \caption{Average inference times of planners.}
    \begin{tabular}{|c|c|c|c|c|c|}
        \hline
        \textbf{Task}  & \makecell{\textbf{Episode} \\ \textbf{Length}} & \textbf{Planner} & \makecell{\textbf{Num.} \\ \textbf{Iter.}} & \makecell{\textbf{Planning} \\ \textbf{Horizon}} & \makecell{\textbf{Inference} \\ \textbf{Time}} \\ \hline
        \texttt{Walk} & 8.0s & iLQG & 10 & 0.35s & 1484.33s \\ \hline
        \texttt{Stand} & 2.0s & iLQG  & 2 & 0.35s & 55.16s \\ \hline
        \texttt{Push} & 1.0s & Sampling &  5 & 1.0s & 38.93s \\ \hline
    \end{tabular}
    \label{table:inference_times}
    \vspace{-1em}
\end{table}
While cheap at train-time (no pre-training needed), MPC is expensive at test-time because of online re-planning.
Therefore, the choice of the i) planner, ii) planning horizon, and iii) number of planning iterations is crucial.
Table~\ref{table:inference_times} shows the run times of our experiments on a MacBook Air M1, 16GB, averaged over 12 runs.
Each planner was given a fixed number of iterations to ensure a consistent comparison (normally, the planner would run as many iterations as possible within a real-time simulation timestep).

\section{Conclusion}

In this paper, we made a step towards enabling easy experimentation with MPC for humanoid robot control in simulation.
We ported \texttt{HumanoidBench} to MuJoCo MPC and provided evaluations on a subset of tasks.
We showed that the current reward functions are insufficient and proposed shaping terms.
Furthermore, we identified an issue with short episode lengths and argued for evaluating on repeated tasks with changing goals.
Our results show superior performance of our method over MPC and RL baselines.

\bibliographystyle{plain}
\bibliography{references} 

\begin{thebibliography}{10}

\bibitem{haarnoja2018soft}
Tuomas Haarnoja, Aurick Zhou, Pieter Abbeel, and Sergey Levine.
\newblock Soft actor-critic: Off-policy maximum entropy deep reinforcement learning with a stochastic actor.
\newblock In {\em International conference on machine learning}, pages 1861--1870. PMLR, 2018.

\bibitem{hafner2023mastering}
Danijar Hafner, Jurgis Pasukonis, Jimmy Ba, and Timothy Lillicrap.
\newblock Mastering diverse domains through world models.
\newblock {\em arXiv preprint arXiv:2301.04104}, 2023.

\bibitem{hansen2023td}
Nicklas Hansen, Hao Su, and Xiaolong Wang.
\newblock Td-mpc2: Scalable, robust world models for continuous control.
\newblock {\em arXiv preprint arXiv:2310.16828}, 2023.

\bibitem{hansen2024hierarchical}
Nicklas Hansen, Jyothir SV, Vlad Sobal, Yann LeCun, Xiaolong Wang, and Hao Su.
\newblock Hierarchical world models as visual whole-body humanoid controllers.
\newblock {\em arXiv preprint arXiv:2405.18418}, 2024.

\bibitem{he2024learning}
Tairan He, Zhengyi Luo, Wenli Xiao, Chong Zhang, Kris Kitani, Changliu Liu, and Guanya Shi.
\newblock Learning human-to-humanoid real-time whole-body teleoperation.
\newblock {\em arXiv e-prints}, pages arXiv--2403, 2024.

\bibitem{howell2022predictive}
Taylor Howell, Nimrod Gileadi, Saran Tunyasuvunakool, Kevin Zakka, Tom Erez, and Yuval Tassa.
\newblock Predictive sampling: Real-time behaviour synthesis with mujoco.
\newblock {\em arXiv preprint arXiv:2212.00541}, 2022.

\bibitem{moerland2023model}
Thomas~M Moerland, Joost Broekens, Aske Plaat, Catholijn~M Jonker, et~al.
\newblock Model-based reinforcement learning: A survey.
\newblock {\em Foundations and Trends{\textregistered} in Machine Learning}, 16(1):1--118, 2023.

\bibitem{schulman2017proximal}
John Schulman, Filip Wolski, Prafulla Dhariwal, Alec Radford, and Oleg Klimov.
\newblock Proximal policy optimization algorithms.
\newblock {\em arXiv preprint arXiv:1707.06347}, 2017.

\bibitem{sferrazza2024humanoidbench}
Carmelo Sferrazza, Dun-Ming Huang, Xingyu Lin, Youngwoon Lee, and Pieter Abbeel.
\newblock Humanoidbench: Simulated humanoid benchmark for whole-body locomotion and manipulation.
\newblock {\em arXiv preprint arXiv:2403.10506}, 2024.

\bibitem{tassa2012synthesis}
Yuval Tassa, Tom Erez, and Emanuel Todorov.
\newblock Synthesis and stabilization of complex behaviors through online trajectory optimization.
\newblock In {\em 2012 IEEE/RSJ International Conference on Intelligent Robots and Systems}, pages 4906--4913. IEEE, 2012.

\bibitem{todorov2012mujoco}
Emanuel Todorov, Tom Erez, and Yuval Tassa.
\newblock Mujoco: A physics engine for model-based control.
\newblock In {\em 2012 IEEE/RSJ international conference on intelligent robots and systems}, pages 5026--5033. IEEE, 2012.

\end{thebibliography}

\end{document}